# Process, Structure, and Modularity in Reasoning with Uncertainty

Bruce D'Ambrosio
Department of Computer Science
Oregon State University

## Abstract

A computational approach to uncertainty management requires support for interactive and incremental problem formulation, inference, hypothesis testing, and decision making. Most current uncertainty inference systems concentrate primarily on inference, and provide no support for the larger issues. We present a computational approach to uncertainty management which provides direct support for the dynamic, incremental aspect of this task, while at the same time permitting direct representation of the structure of evidential relationships. At the same time, we show that this approach responds to the modularity concerns of Heckerman and Horvitz [Heckerman & Horvitz, 1987]. This paper emphasizes examples of the capabilities of this approach. Another paper [D'Ambrosio, 1988] details the representations and algorithms involved.

## 1 Introduction

A review of the literature on uncertainty in AI might lead one to conclude that the problem consists solely of choosing an appropriate certainty calculus. However, a complete approach to uncertainty management requires support for interactive and incremental problem formulation, inference, hypothesis ranking, and decision making. Further, it should be based on a normative model of reasoning and decision-making under uncertainty, must provide support for defeasible decision-making about problem model formulation, and must offer ways of bounding the resources need for uncertainty management.

In this paper we present an approach which begins to address these requirements. It is based on the belief that while Bayesian inference provides one possible "gold standard" for micro-management of uncertainty, that is, describing and evaluating isolated instances of uncertainty reasoning, it ignores the macro-management aspects of reasoning with uncertainty, the problems of how an intelligent agent goes about structuring and revising situation models. In our approach, uncertainty is represented symbolically and structurally. The approach permits natural and intuitive representations of commonly occurring evidential structures, permits analysis even in the face of weak, non-numeric information about beliefs, and yet conforms to a normative model of inference when full probability data is available. At the same time, the symbolic underpinnings of our approach also provide a mechanism for incremental construction, modification, and evaluation of belief networks. We believe this to be an essential aspect of uncertainty reasoning which has been ignored in most previous research. In addition, they provide a natural mechanism for modelling macro-management of uncertainty reasoning as the dynamic construction and evaluation of a set of related situation models, based on a subgoalling structure. In subsequent sections of this paper we first briefly introduce Hybrid Uncertainty Management (HUM), and then illustrate its operation by considering two examples which have appeared recently in the literature on uncertainty in AI.

## 2 Related Research

Breese [Breese, 1987] has studied the problem of model formulation, starting from a database containing both logical and probabilistic knowledge. Laskey [Laskey, 1987] has independently begun developing a mapping for probabilistic knowledge into an ATMS similar to ours. Quinlan [Quinlan, 1987] has considered the problems of hard-wiring assumptions regarding model structure, and resulting limitations in expressivity. Henrion [Henrion, 1987], Pearl [Pearl, 1987], Shachter [Shachter & Heckerman, 1987], and others are studying performance models with the full expressivity of probability theory.

## 3 Hybrid Uncertainty Management

### 3.1 Issues and Overview

Hybrid uncertainty management is a framework for reasoning with uncertain information based on the following principles:

1. All information, including information about our uncertainty, should be represented using *explicit, local* representations. These representations should preserve the *structure* of information, as well as its "content".

2. For efficient yet robust reasoning, representations must be such that they can be incrementally constructed, evaluated, and modified, and the same uncertainty management capabilities should be available to the agent when reasoning about construction of a model, as when reasoning about the domain problem.

We are constructing[1] an Uncertainty Inference System (UIS [Henrion, 1987]) in accordance with these principles, based on the logical inference capabilities of an Assumption-Based Truth Maintenance System (ATMS) [deKleer, 1986]. An ATMS provides a complete propositional logic. That is, given a set of atomic propositions and a set of logical formulas over those propositions, it computes the set of consistent assignments of truth values to a core set of propositions, and derives the truth value of every other proposition in terms of the truth values of the core set. Derived truth values are represented as expressions, in disjunctive normal form, for the truth of every proposition in terms of the truth of core propositions. It is this DNF expression, called a *label* in ATMS parlance, which provides the core structured representation for uncertainty in HUM.

In creating HUM, we are developing mechanisms for using the representational and inferential capabilities of an ATMS for both micro- and macro-management of reasoning with uncertainty. For micro-management, we have added mechanisms which permit the interpretation of selected *assumptions* as markers for elements of *probability distributions*, and have developed mappings for the basic expressions in a probabilistic model. This required developing the following components:

1. A representation for a probability distribution - we use the ATMS *choose* operator to indicate that a set of assumptions represent a probability distribution, and annotate each assumption with the corresponding numeric probability, when available.

2. A mapping for the basic computations of probabilistic inference - we use the ATMS *justification* as the primitive component of our representation for an probabilistic relationship between variables, and the ATMS *environment propagation* mechanism to perform inference.

---
[1] All the examples shown run in the current prototype.

3. A mechanism for retrieving probabilities from ATMS labels - we have developed an algorithm which can evaluate labels on request to reduce them to a numeric representation.

For further details see [D'Ambrosio, 1987b], [D'Ambrosio, 1988].

Macro-management requires the ability to:

1. Represent and reason about the decisions involved in model construction.

2. Recognize when inference at the domain level indicates problems in the model formulation, and revise domain models as appropriate.

The domain models are constructed as ATMS networks, and ATMS *assumptions* are used to explicitly record defeasible model structure decisions. As we will show in an example below, this permits dynamic extension and revision, as well as incremental evaluation, of domain models. Our current implementation provides a mixed forward and backward chaining rule language in which model construction algorithms are written. Forward chaining is used to express the basic model construction algorithms, and backward chaining is invoked when information needed during model construction is unavailable. This architecture provides the same capabilities in response to model structure subgoals as are available at the domain level. We illustrate in a second example how explicit recording of structuring decisions permits revision and incremental re-evaluation of domain models.

### 3.2 Example One: The Three Urns

We begin with a very simple example, the three urns problem described in [Heckerman & Horvitz, 1987]:

> Suppose you are given one of three opaque jars containing mixtures of black licorice and white peppermint jelly beans. The first jar contains one white jelly bean and one black jelly bean, the second jar contains two white jelly beans, and the third jar contains two black jelly beans. You are not allowed to look inside the jar, but you are allowed to draw beans from the jar, *with* replacement. That is, you must replace each jelly bean you draw before sampling another. Let $H_i$ be the hypothesis that you are holding the *ith* jar. As you are told that the jars were selected at random, you believe that each $H_i$ is equally likely before you begin to draw jelly beans.

We can represent this situation in HUM in the following fashion[2]:

```
;;; I have one of three possible urns
(Variable Urn H1 H2 H3)
;; every draw has two possible outcomes
(Variable (Draw ?n) white black)
```

The expression (Draw ?n) above states that we are describing, not a single draw, but a class of possible draws from an urn. Actual Draws will instantiate the logical variable ?n with a number 1 .... We can express the conditional probability of drawing a white based on the urn we are holding as follows:

---

[2]In the following psuedo-lisp examples, we simply present an expression when the value returned is not of interest. If the returned value is relevant, we precede the expression by > and show the result on the following line.

```
;;; probability of drawing white, for each urn:
(Relation Urn (Draw ?n)
        (-> (Urn H1) (((Draw ?n) white) .5)(((Draw ?n) black) .5))
        (-> (Urn H2) (((Draw ?n) white) 1.0))
        (-> (Urn H3) (((Draw ?n) white) 0.0)))
```

Each "->" in this example corresponds to a rule. Finally, we can express our pre-existing information about jars as follows:

(Marginal Urn (Urn H1) .33 (Urn H2) .33 (Urn H3) .33)

### 3.2.1 Incremental Symbolic Evaluation

The above information serves to fully constrain the conditional probabilities for a class of possible situation models. Note, however, that the full joint distribution need never be explicitly represented, either in the problem statement or in the ATMS network: the label reduction algorithm can directly combine the marginals for each evidence source. At this point, the specific situation model constructed so far only contains the $Urn$ variable. We can now ask HUM for the probability that we are holding H2:

```
>(Probability-of (Urn H2))
0.33
```

We can also ask for the probability of selecting a white jelly bean, by first extending the model to include a first draw, and then querying the probability of various outcomes:

```
(Instance (draw 1))
>(Probability-of ((draw 1) white))
.5
```

Any of this could easily be done in several of the various UIS's currently available, and should seem rather boring (except, perhaps, for the ability to describe a *class* of variables, (Draw ?n). Consider, however, what happens once we actually draw a sample:

```
(Deffactq ((draw 1) white))
>(Probability-of (Urn H2))
0.67

(Instance (draw 2))
(Deffactq ((draw 2) white))

>(Probability-of (Urn H2))
0.8
```

The system *incrementally* computes the new posterior for the various Urn hypotheses, by incrementally extending the network and incrementally updating the structured representation for the probability of each urn.

A fragment of the ATMS network corresponding to this problem is shown in figure 1. Arcs in the diagram represent ATMS justifications, in this case used to express conditional probability relationships between variables. The label of each value is shown in brackets under the value.

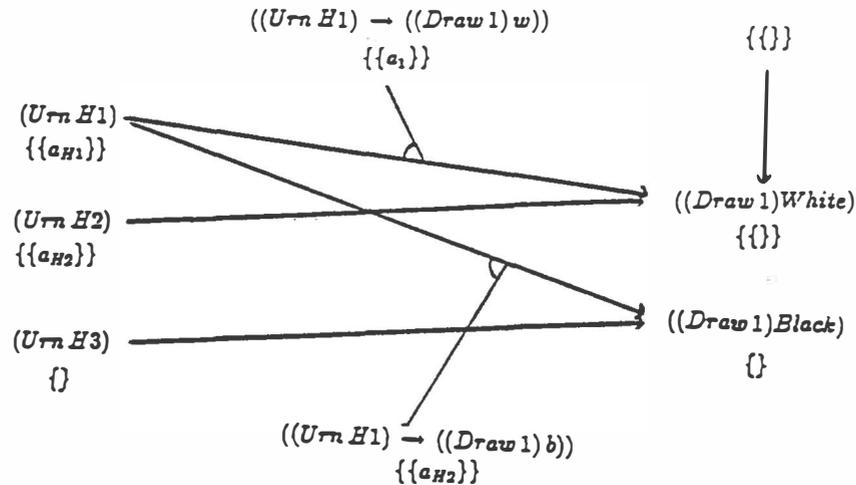

Figure 1: ATMS network for Urn Problem after Draw 1

Thus, the ATMS records that hypothesis (Urn H2) is true precisely under the assumption $a_{H2}$, wich carries the original marginal assigned, while $((Draw1)white)$ is true universally (that is, it is true in a environment which includes no assumptions). $((Draw1)black)$, on the other hand, is not supported in any environment. It may not be obvious from this example how the observation of draw 1 affects the probability of holding Urn H2, since the label for Urn H2 is unchanged. This takes place through the *nogood* database, an ATMS maintained data structure containing a minimal representation of all environments shown to be invalid. When $((Draw1)white)$ is asserted, the environments in the label of $((Draw1)black)$ become nogood. Since these include assumptions used to represent the marginal for Urns, the subsequently computed probability for any value of the Urns variable is changed.

A complete computational treatment of uncertainty must meet a variety of requirements. Some are expressivity requirements, as identified by Heckerman and Horvitz [Heckerman & Horvitz, 1987]. Others arise from the dynamic nature of computation and interaction with the world. We have used this example to illustrate how several of these requirements are handled in HUM. We have shown how HUM handles mutual exclusivity, bidirectional inference (reasoning from causes to observables and observables to causes interchangeably), and one process aspect of reasoning with uncertainty, incremental construction and evaluation of models. Note also that this representation system does not suffer from the modularity problems observed by Heckerman and Horvitz when they considered the same example. The "rules" relating urns to outcomes need only be expressed once, and remain correct for any number of draws and any sequence of outcomes.

### 3.3 Example 2 - Thousands Dead

A major issue in uncertainty reasoning is the representation and use of correlated evidence. In this next example, taken from [Henrion, 1987], we examine reasoning about model structure and model revision. Specifically, we attempt to show that commonly occurring correlations between evidence sources are the result of *structural* relationships between such sources, and therefore are best supported by a system which permits direct expression of that structure:

> **Chernobyl example:** The first radio news bulletin you hear on the accident at the Chernobyl nuclear power plant reports that the release of radioactive materials may have already killed several thousand people. Initially you place small credence in this, but as you start to hear similar reports from other radio and TV stations, and
in the newspapers, you believe it more strongly. A couple of days later, you discover
that the news reports were all based on the same wire-service report based on a single
unconfirmed telephone interview from Moscow. Consequently, you greatly reduce your
degree of belief again.

We see two interesting issues here. The first is expressivity. Henrion points out that "...none of the better known UISs are actually capable of distinguishing between independent and correlated sources of evidence." One view is that the problem here is not probabilistic, but rather a logical uncertainty about the possible coreference of evidence for the various reports. The second is a process issue. What changes is our beliefs about the *structure* of the evidential relationships. We are unaware of any existing UIS that is capable of accommodating this structural change. All require reconstructing the model in its entirety, which we claim is exorbitant and unrealistic.

The information we have underconstrains the possible joint evidential relationship between news reports and the number dead. One possibility, advocated by Quinlan [Quinlan, 1987], is to abstain from committing, and compute a weak, interval-valued result. We believe this is unrealistic. Commitments must be made in the absence of conclusive data, based on *reasonable assumptions* about the situation being modelled. What is crucial is that a mechanism be provided through which an agent can reason about and record structural assumptions. The structured uncertainty representation provided by an ATMS provides the bookkeeping necessary for such a mechanism, as we now show.

Our initial knowledge is as follows:

```
    ;;;thousands of people died
 (Variable 1000s-dead true false)
 ;;;I heard in on the radio or in the Newspaper
 (Variable (Radio ?n) true false)
 (Variable (News ?n) true false)

 (Relation (Radio ?n) 1000s-dead
       (-> (((Radio ?n) true)) (((1000s-dead true) 1.0)))
 (Relation (News ?n) 1000s-dead
       (-> (((News ?n) true)) (((1000s-dead true) 1.0)))
```
;;;we know something about the source of news
;;; upi, associated press, or independent
(Variable (source (radio ?n)) elem upi ap ind)
(Variable (source (news ?n)) elem upi ap ind)

If we now assert the receipt of a radio report, we can compute the belief in thousands dead.

```
    (Instance (radio 1))
 (Instance (source (radio 1)))
 ;;; we tend to believe radio reports
 (Deffactq (Marginal (radio 1) .7 .3))
 ;;; we have no information regarding source of info
 (Deffactq (marginal (source (radio 1)) (.33 .33 .34)))
 >(Probability-of (1000s-dead true))
 0.7
```

However, if we subsequently assert the receipt of a second radio report, or a newpaper report, the joint distribution is underconstrained, and the posterior belief in thousands-dead cannot be computed. if we assume that each report provides *independent* evidence, then we have enough information to recover the complete joint distribution. Now what happens when we change our mind about the independence of the various reports? First we show system operation, then explain what happened:

```
      (Instance (news 1))
   (Instance (source (news 1)))
   (Deffactq (Marginal (source (news 1)) (.33 .33 .34)))
   (Deffactq (Marginal (news 1) (.7 .3)))
   ** Assuming (Independent evidence-for (radio 1) (news 1)) ***
   ** Monitoring (Same evidence-for (radio 1) (news 1)) ***
   >(Probability-of (1000s-dead true))
   0.91
   (Deffactq (Same evidence-for (radio 1) (news 1)))
   ** Retracting (Independent evidence-for (radio 1) (news 1)) ***
   >(Probability (1000s-dead true))
   0.7
```

When reasoning about model structure, the system has all of the reasoning capabilities available at the domain model level. When it detects an attempt to instantiate a second relationship with (1000s-dead) as the consequent variable, it establishes a subgoal to determine the relationship between the two sets of antecedents. This subgoal is solved by constructing simple decision model, based on information provided above about evidence sources, which decides how to instantiate the evidence for the second report. In our example, the prior probability that sources are independent is high enough to result in the decision to instantiate the evidence for the second report as a separate, independent source of evidence. However, because the evidence for independence is not conclusive, a *monitor* is installed on the possibility that the evidence sources are the same. When evidence later arrives that the sources of evidence for the two reports are in fact identical, this triggers the monitor to re-structure the domain model in accordance with the new information. This restructuring is accomplished by invalidating the previous evidence for (news 1) and sharing the evidence source for (radio 1) between the two reports. As we stated earlier, the assumption of independence is the statement of a *lack* of structural connection between the various reports. In order to make this assumption retractable, we condition the connection between the original evidence for (news 1) and (news 1) on the independence assumption. Thus, if the independence assumption is $a_2$, the above described situation is represented by the following ATMS network:

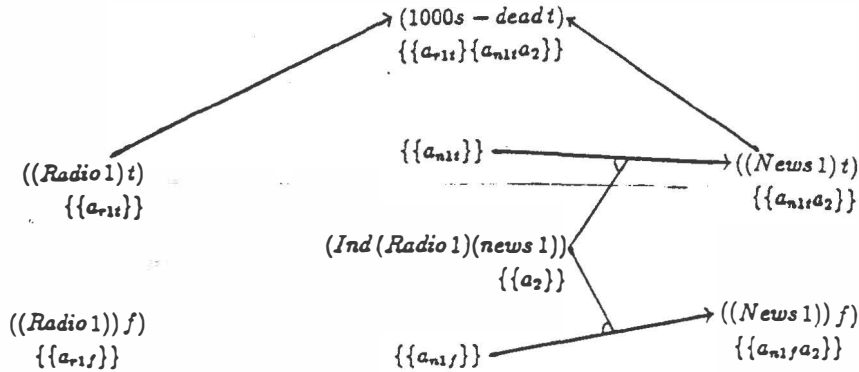

The purpose for explicitly recording the independence assumption is to permit later retraction. When we later discover that the evidence for the reports is not independent, we retract the independence assumption, and extend the model to reflect our updated understanding of the situation (we omit some of the old arcs for clarity):

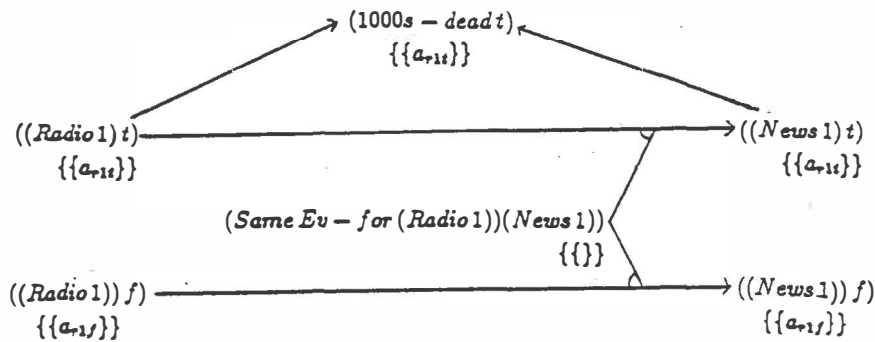

## 3.4 Other Issues

Various numeric uncertainty formalisms each have fervent adherents. We are more interested in the *structure* of such reasoning. In an attempt to understand the impact of the particular formalism chosen on the process of reasoning under uncertainty, we are developing mappings similar to the one shown, based on both belief functions and fuzzy set theory. While we believe that we have developed representations capable of capturing any arbitrary probabilistic relationships among a set of variables (see [D'Ambrosio, 1988]), we have barely begun to incorporate the machinery of decision analysis into this framework. Also, our current mechanisms for control of the reasoning process are crude and only handle a few special cases. We are incorporating our approach into a blackboard-like architecture which will provide more flexibility in control of uncertain inference. The structured symbolic representation provides fertile ground for exploring the problem of explanation of results, a topic we have not yet begun to explore. Finally, our current implementation is an early prototype, and its speed is marginally adequate for the experiments we would like to run.

## 4 Summary

We have presented an approach to uncertain inference which emphasizes both expressivity and process. We identify two aspects of expressivity: adequacy and economy. We have illustrated through examples that this approach is not only adequately expressive, but also economical, in that it captures *structurally* commonly occurring classes of evidential relationships. Equally critical, we emphasize the need to support the *process* of uncertain inference, that is, the incremental construction and evaluation of probabilistic domain models.